\documentclass[final]{cvpr}

\usepackage{times}
\usepackage{epsfig}
\usepackage{graphicx}
\usepackage{amsmath}
\usepackage{amssymb}
\usepackage{booktabs}
\usepackage{float}
\usepackage{bm}
\usepackage{booktabs}
\usepackage{multirow}
\usepackage{makecell}
\usepackage{color}
\usepackage{algorithm}
\usepackage{algorithmicx}

\usepackage{algpseudocode}
\usepackage{subfigure}
\usepackage{CJK}

\usepackage[pagebackref=true,breaklinks=true,colorlinks,bookmarks=false]{hyperref}

\def\cvprPaperID{2369} 
\def\confYear{CVPR 2021}

\begin{document}
	
	\title{Multi-scale Adaptive Task Attention Network for Few-Shot Learning}
	
	\author{Haoxing Chen, Huaxiong Li, Yaohui Li, Chunlin Chen\\
		Nanjing University\\
		{\tt\small $\lbrace$haoxingchen, yaohuili$\rbrace$@smail.nju.edu.cn, $\lbrace$huaxiongli, clchen$\rbrace$@nju.edu.cn}
	}
	
	\maketitle

	\begin{abstract}
The goal of few-shot learning is to classify unseen categories with few labeled samples. Recently, the low-level information metric-learning based methods have achieved satisfying performance, since local representations (LRs) are more consistent between seen and unseen classes. However, most of these methods deal with each category in the support set independently, which is not sufficient to measure the relation between features, especially in a certain task.  Moreover, the low-level information-based metric learning method suffers when dominant objects of different scales exist in a complex background. To address these issues, this paper proposes a novel Multi-scale Adaptive Task Attention Network (MATANet) for few-shot learning. Specifically, we first use a multi-scale feature generator to generate multiple features at different scales. Then, an adaptive task attention module is proposed to select the most important LRs among the entire task. Afterwards, a similarity-to-class module and a fusion layer are utilized to calculate a joint multi-scale similarity between the query image and the support set. Extensive experiments on popular benchmarks clearly show the effectiveness of the proposed MATANet compared with state-of-the-art methods. 
	\end{abstract}
	
	
	\section{Introduction}
	
	
	\begin{figure}
		\centering          
		\subfigure[]{       \begin{minipage}{8.5cm}      \centering      \includegraphics[height = 6cm, width=8.5cm]{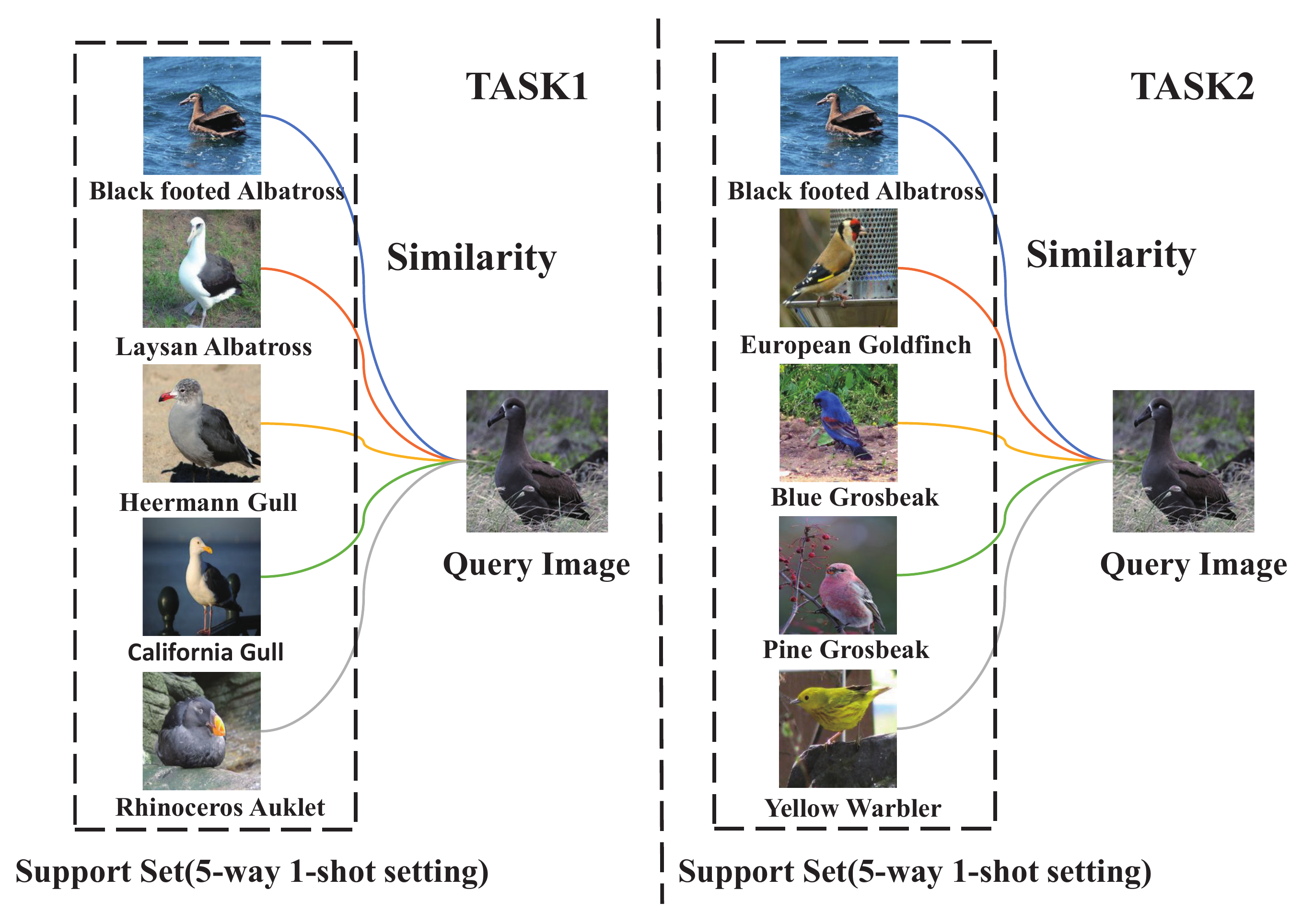}      \end{minipage}      }      
		\subfigure[]{       \begin{minipage}{7cm}      \centering      
				\includegraphics[height = 2.5cm, width = 7 cm]{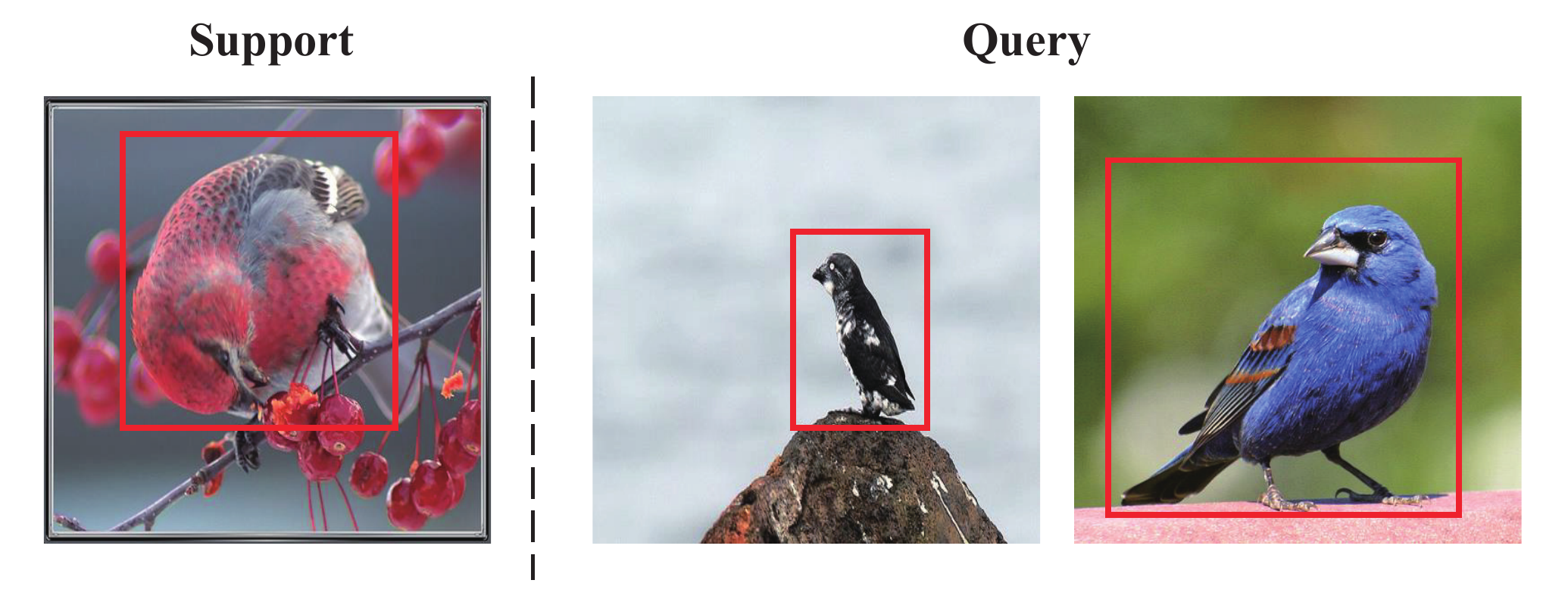}      \end{minipage}      }      \caption{The two main problems of the previous local representation based methods. (a) In different tasks, the most discriminative features are different. In task 1, the beak is the key distinguishing feature, while the most critical feature is the wing in task 2.
			(b) The scales of the dominant objects varies from image to image.}      \label{fig1}  
	\end{figure}
	Humans can learn new concepts and objects with only one or a few samples easily.
	In order to imitate this ability of humans, many few-shot learning methods \cite{vinyals2016matching,finn2017model,snell2017prototypical,sung2018learning,sun2019meta,li2019revisiting,li2019distribution} have been proposed. However, most of these methods \cite{vinyals2016matching,finn2017model,snell2017prototypical,sung2018learning,sun2019meta} adopt image-level features for classification.
	Due to the scarcity of samples in few-shot image recognition tasks, classifying at such a level may not be effective enough. Instead, many methods \cite{li2019revisiting,li2019distribution} based on low-level information were proposed, i.e., local representations (LRs) of feature embeddings. 
	These methods use low-level information to measure the distance between query images and support images, and they can achieve better recognition results.
	However, these methods do not measure the similarity between query images and support images in the context of the whole task, which does not make full use of the representation ability of local feature descriptors. When humans classify an image into one of several unseen classes, it is natural to look for semantic features that are shared only between certain classes and the query image. In other words, humans do not pay much attention to the features shared between classes when recognizing a category they have not seen before. 
	For example, consider two 5-way 1-shot tasks in Figure 1(a). In task1, we need to recognize a `Black-footed Albatross' among `Laysan Albatross', `Heermann Gull', `California Gull' and `Rhinoceros Auklet'. While in task 2, we need to recognize a `Black-footed Albatross' among `European Goldfinch', `Blue Grosbeak', `Pine Grosbeak' and `Yellow Warbler'. For task 1, the beak is a very distinguishing feature, but for task 2 it is not the most critical feature. Similarly, the wing is more important for task 2 than task 1. In summary, the importance of each LRs varies from task to task.
	
	As previously mentioned, although these existing methods can extract the relation between the query image and each support set independently, they do not consider the importance of each LRs under the whole task, and all the LRs are weighted equally, rather than the task-relevant LRs enjoy the higher weights. Moreover, these methods can only calculate their similarity at a single scale. As shown in Figure 1(b), due to the scales of dominant objects in different images are dissimilar, we argue that it is more reasonable to calculate the similarity between query image and support set at multiple scales simultaneously.
	
	To this end, we propose a novel \emph{Multi-scale Adaptive Task Attention Network} for metric-learning based few-shot learning, which can be trained in an end-to-end manner. First, we represent all images as a collection of LRs at different scales by a multi-scale feature generator, rather than a global feature representation at the image level. Second, we measure the semantic similarity between the query image and the support set by calculating the semantic relation matrix. Afterwards, we employ an adaptive task attention module to select the most distinguishing feature in the current task. Third, to further make full use of LRs, we employ a similarity-to-class mechanism to determine which support class the query image belongs to at each scale. Finally, we adaptively fuse the similarities calculated from the features of different scales together.
	
	 To sum up, the main contributions are summarized as follows:
	\begin{itemize}
		\setlength{\itemsep}{0pt}
		\setlength{\parsep}{0pt}
		\setlength{\parskip}{0pt}
		\item To generate different scales features, we propose a multi-scale feature generator in few-shot learning tasks, which can provide multi-scale information for more comprehensive measurements.
		\item We further propose a novel adaptive task attention mechanism by finding and weighing most discriminative local representations in the entire task, aiming to learn task-relevant feature representations for few-shot learning.
		\item We conduct sufficient experiments on four benchmark datasets to verify the advancement of our model, and the performance of our model achieves the state-of-the-art.
	\end{itemize}

	\section{Related Work}
	In the recent few-shot learning literature, there are roughly two types of methods: meta-learning based method and metric-learning based method.
	
	\textbf{Meta-learning based methods.} 
	The main idea of meta-learning based methods \cite{ravi2016optimization,finn2017model,sun2019meta,jamal2019task} is how to use existing experience or knowledge reasonably to achieve fast learning when faced with a new task, rather than starting from scratch.
	Santoro et al. \cite{santoro2016meta} adopted an LSTM to control the interaction between the network and the external memory module. 
	Santoro et al. \cite{ravi2016optimization} describe a new meta-learning method by interpreting SGD update rules as a recursive gated model with trainable parameters.
	The purpose of MAML \cite{finn2017model} is to learn a good parameter initialization so that the model can quickly adapt to new tasks.
	Sun et al. \cite{sun2019meta} proposed meta-transfer learning, which learns the scaling and shifting functions of DNN weight for each task.
	Jamel et al. \cite{jamal2019task} proposed a task unbiased method, by introducing regularized loss terms to constrain the model to have as little preference for all tasks as possible when parameters are updated.
	Although these meta-learning based methods have achieved outstanding results on few-shot learning tasks, their complex memory addressing architecture is difficult to train. Compared with these methods, our proposed MATANet can be trained in an end-to-end manner. 
	\begin{figure*}[t]
		\centering
		\includegraphics[height=8cm,width=17.5cm]{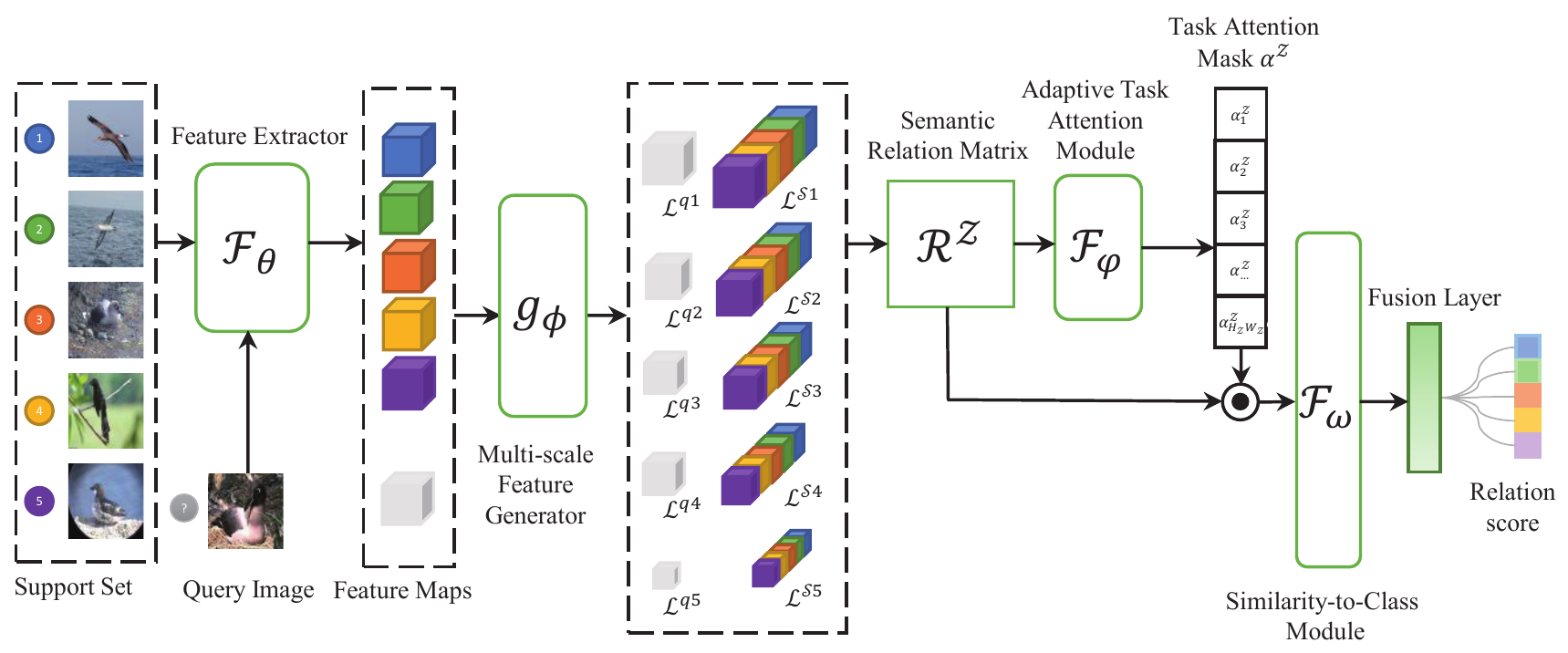}
		\caption{The framework of MATANet under the 5-way 1-shot image classification setting. The model mainly consists of four modules: the feature extractor $\mathcal{F}_{\theta}$ to learn local representations, the multi-scale feature generator $g_{\phi}$ to generate multiple features at different scales, the adaptive task attention module $\mathcal{F}_{\varphi}$ to generate adaptive task attention mask for selecting more important elements of semantic relation matrix, and the similarity-to-class module $\mathcal{F}_{\omega }$ to get a similarity score to determine which support class the query image belongs to. The black square indicates the predicted label. (Best view in color.)}
		\label{atanet_architecture}
	\end{figure*}
	
	\textbf{Metric-learning based methods. }
	The metric-learning based methods aim to learn an informative distance metric, as demonstrated by \cite{vinyals2016matching,snell2017prototypical,sung2018learning,allen2019infinite,garcia2017few,li2019distribution,li2019revisiting,simon2020adaptive}.
	Koch et al. \cite{koch2015siamese} used a Siamese Neural Network to tackle the one-shot learning problem. Snell et al. \cite{snell2017prototypical} proposed Prototypical Networks, which first assumes that each type can be represented by a prototype, and the prototype can be obtained by calculating the mean value of the embedding representation of each class and then using a distance function to classify. In fact, we don't know which distance function is the best. Therefore, Sung et al. \cite{sung2018learning} proposed a Relation Network to obtain the most suitable distance metric function through learning. 
	The above methods are all based on the feature representation at the image level. Due to the scarcity of the number of samples, we can not well represent the distribution of each category on the image level features. In contrast, some recent work, such as DN4 \cite{li2019revisiting} and CovaMNet \cite{li2019distribution} shows that the rich low-level features (i.e., LRs) have better representation capabilities. However, these methods measure the similarity between the query image and each support class independently, without considering the entire task together. Moreover, the methods based on low-level features only measure the similarity between query image and support set at a single scale, which may lead to a lower classification accuracy when the scales of dominant objects are different. 
	
	Unlike the above method, our MATANet calculates the similarity between the query image and the support set at multi-scale. And then we can obtain the final result through integrate multiple similarities from different scales. In addition, our MATANet can adaptively select task-relevant local features with discriminative semantics, as the process of human recognition.

	\section{The proposed Method}
	
	\subsection{Description on Few-shot Learning}
	In few-shot learning, there are usually three sets of data: a query set $\mathcal{Q}$, a support set $\mathcal{S}$, and an auxiliary set $\mathcal{A}$. Note that $\mathcal{Q}$ and $\mathcal{S}$ share the same label space, while they have no intersection with the label space of $\mathcal{A}$. 
	
	In this paper, we follow the definition of the common few-shot learning task.
	Given a support set that contains N previously unseen classes, with K samples for each class.
	We need to determine which class the query sample belongs to, which is called N-way K-shot tasks (e.g., 5-way 1-shot or 5-way 5-shot).
	To achieve this goal, we use an auxiliary set to train a model to learn transferable knowledge. The model is trained by the episodic training mechanism \cite{vinyals2016matching}. In each episode, a new task is constructed randomly in $\mathcal{A}$, and each task consists of two subsets: auxiliary support set $\mathcal{A_S}$ and auxiliary query set $\mathcal{A_Q}$. Generally, in the training stage, hundreds of tasks are adopted to train the model.
	
	As shown in Figure 2, our MATANet is mainly composed of four modules: a feature extractor $\mathcal{F}_{\theta}$, a multi-scale feature generator $g_{\phi}$, an adaptive task attention module $\mathcal{F}_{\varphi}$, and a similarity-to-class module $\mathcal{F}_{\omega }$. 
	All image samples are first fed into the $\mathcal{F}_{\theta}$ to get feature embeddings and rich LRs. In practice the feature extractor module could be 4-layer CNN, ResNet12 \cite{he2016deep} or WRN \cite{zagoruyko2016wide}. Then the multi-scale feature generator generates multiple features at different scales. Afterwards, semantic relation matrixes are calculated to measure the semantic relevance between query image and support set at each scale. The adaptive task attention module learns a task attention mask which can adaptively calculate the importance of each LR in the current task. And we use task attention masks to weighting semantic relation matrix to prominently display task-relevant elements. After that, the weighted semantic relation matrix is processed by similarity-to-class module $\mathcal{F}_{\omega }$ to determine which support class the query image belongs to. Finally, we adaptively fuse the similarities calculated from the features of different scales together by a learned vector. All the modules can be trained jointly in an end-to-end manner.
	\subsection{Multi-scale Feature Generator}
	As some recent studies \cite{li2019revisiting,li2019distribution} on few-shot learning have proved, LRs show richer representation ability and can alleviate the problem of sample scarcity in few-shot learning. Therefore we use LRs to represent the features of each image. Given a query image $\mathcal{A}^q_\mathcal{Q}$ through the feature extractor we can get a three-dimensional (3D) vector $\mathcal{F}_{\theta}(\mathcal{A}^q_\mathcal{Q})\in\mathbb{R}^{C\times H\times W}$. Under the  N-way K-shot few-shot learning setting, there are K images for each support class in a certain task. Through feature extractor we can get a four-dimensional (4D) vector of support set $\mathcal{S}$, which can be denoted as $\mathcal{F}_{\theta}(\mathcal{S})\in\mathbb{R}^{NK\times C\times H\times W}$.
	
	\begin{figure}[t]
		\centering
		\includegraphics[height=6cm,width=8cm]{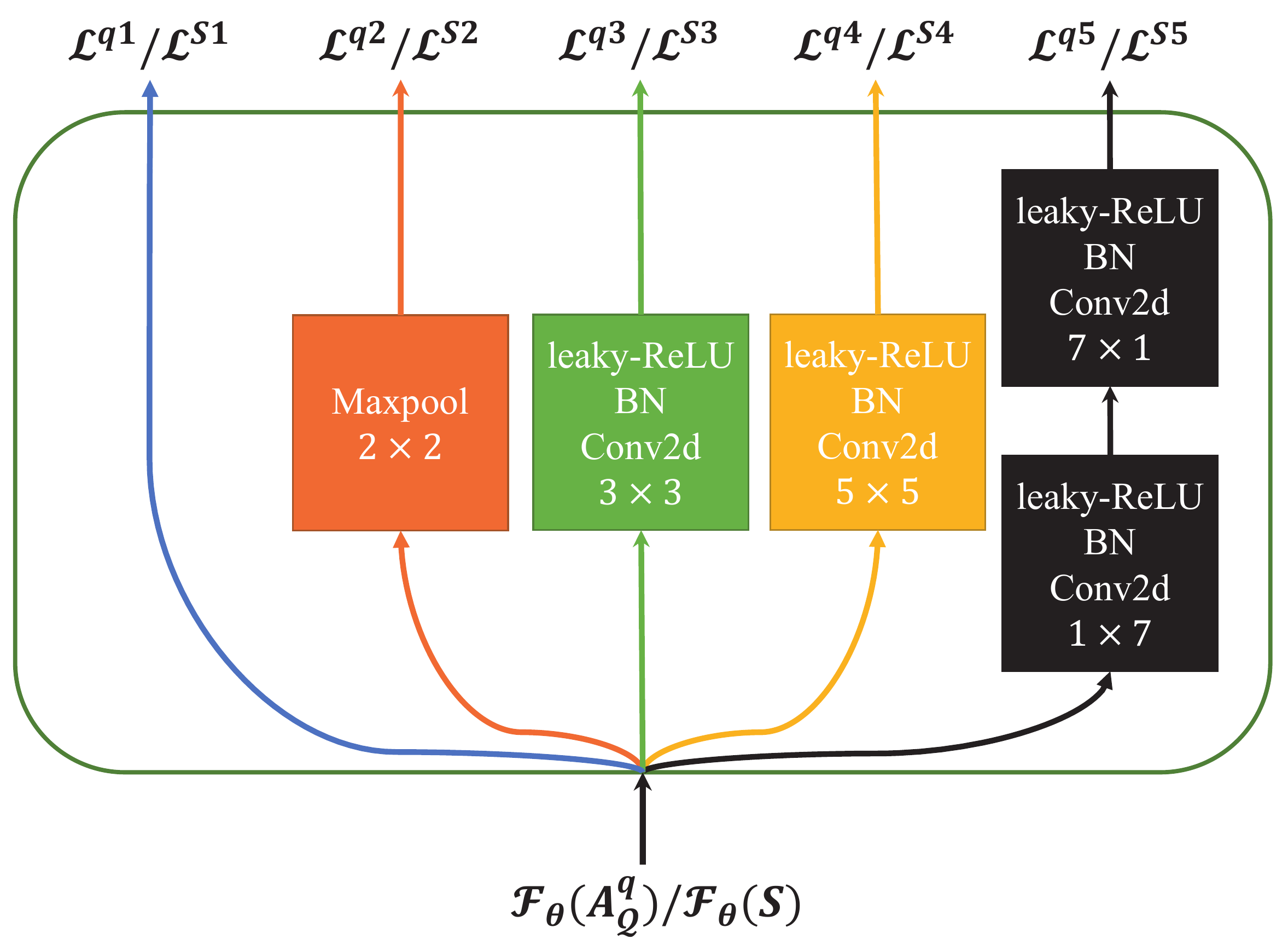}
		\caption{The architecture of the multi-scale feature generator $g_{\phi}$. }
		\label{generator}
	\end{figure}
	
	The multi-scale feature generator aims to generate multiple features at different scales. As illustrated in Figure 3, the multi-scale feature generator consists of five components: the first part has no processing unit, the input and output are the same; the second part has a $2\times2$ max-pooling layer; the third part has a $3\times3$ convolutional layer; the fourth part has a $5\times5$ convolutional layer; the fifth part has a $1\times7$ convolutional layer and a $7\times1$ convolutional layer. 
	
	Through the multi-scale feature generator, we can get the 3D features  $\mathcal{L}^{qz}\in\mathbb{R}^{C_z\times H_z\times W_z} ,z\in\left\{1,2,3,4,5\right\}$, which can be regarded as a set of $H_z\times W_z$ $C_z$-dimensional LRs
	\begin{equation} 
		\mathcal{L}^{qz}=[x_1,...,x_{H_zW_z}]\in\mathbb{R}^{C_z\times H_zW_z} 
	\end{equation}
	where $x_{i}$ is the $i$-th LRs. Through multi-scale feature generator we can also get the LRs of support set $\mathcal{S}$ 
	\begin{equation} 
		\mathcal{L}^{\mathcal{S}z}=[x_1,...,x_{NKH_zW_z}]\in\mathbb{R}^{C_z\times NKH_zW_z} 
	\end{equation}
	
	\subsection{Adaptive Task Attention Module}
	Under the N-way K-shot few-shot learning setting, we calculate the semantic relation matrix $\mathcal{R}^z$ between a query image $\mathcal{A}^q_\mathcal{Q}$ and support set $\mathcal{S}$ to measure semantic relevance by LRs at the $z$-th scale. Then the $\mathcal{R}^z$ can be calculated as below
	\begin{gather}
		\mathcal{R}_{i,j}^z = cos(\mathcal{L}^{qz}_i,\mathcal{L}^{\mathcal{S}z}_j)
		\\
		cos(\mathcal{L}^{qz} _i,\mathcal{L}^{\mathcal{S}z}_j) =
		\frac{(\mathcal{L}^{qz} _i)^T\mathcal{L}^{\mathcal{S}z}_j}
		{
			\left\|\mathcal{L}^{qz} _i \right\| \cdot  \left\|\mathcal{L}^{\mathcal{S}z}_j\right\|} 	
	\end{gather}
	where $i\in\left\{1,...,H_zW_z\right\}, j\in\left\{1,...,NKH_zW_z\right\}, z\in\left\{1,2,3,4,5\right\}$,  $\mathcal{R}^z_{i,j}$ is $(i,j)$ element of $\mathcal{R}^z$ reflecting the distance between the $i$-th LR of the query image and the $j$-th LR of support set at the $z$-th scale and $cos (\cdot, \cdot)$ is Cosine  distence function. 
	
	Each row in $\mathcal{R}^z$ represents the semantic relation of each LR in the query image to all LRs of all images in the support set, i.e., semantic relation vector  $\mathcal{R}^z_i$ represent the relation between $i$-th LR of query image $\mathcal{A}^q_\mathcal{Q}$ to all $NKH_zW_z$ LRs of support set at $z$-th scale. $\mathcal{R}^z$ can be decomposed into N submatrices $\mathcal{R}^{zn}$, $n\in\left\{1,...,N\right\}$ according to columns, representing the semantic relation between the query image and each support class. 
	
	Then we can calculate the task attention score of each element of $\mathcal{R}^z$ for the current task by
	\begin{equation} 
		\alpha^z_i = \frac{
			\sum_{j=1}^{NKH_zW_z}\mathcal{R}^z_{i,j}
		}{
			\sqrt{\sum_{i=1}^{H_zW_z}\sum_{j=1}^{NKH_zW_z}\mathcal{R}^z_{i,j} }}
	\end{equation}
	The task attention mask $\alpha^z$  is consist of all task attention scores $\alpha^z_i$, $i\in\left\{1,...,H_zW_z\right\}$. Afterwards, we use dot-product to weight $\mathcal{R}^z_i$ by $\alpha^z_i$
	\begin{equation} 
		\mathcal{M}^z_i =
		\alpha^z_i \cdot \mathcal{R}^z_i
	\end{equation}
	where $\mathcal{M}^z_i$ is the $i$-th row of weighted semantic relation matrix. By this way, we can get the weighted semantic relation matrix $\mathcal{M}^z$, which can be decomposed into N submatrices $\mathcal{M}^{zn}$, $n\in\left\{1,...,N\right\}$ according to columns. While the semantic relations of task-irrelevant regions are suppressed; meanwhile, the semantic relations of task-relevant regions are enhanced. 
	
	\subsection{Similarity-to-Class Module}
	Similarity-to-Class Module aims to determine which support class the query image belongs to. In this module, for each LR of the query image, we find the $k$ most similar LR of all support LRs for class $n$. Then, we sum $kH_zW_z$ selected LRs as the similarity score between the query image and the $n$-th support class at the $z$-th scale
	\begin{equation} 	
		\mathcal{P}^{zn}  = \sum_{i=1}^{kH_zW_z}Topk(\mathcal{M}^{zn}_i)	
	\end{equation}
	where $\mathcal{P}^{zn}$ means the similarity between the query image and support class $n$ at the $z$-th scale and $Topk (\cdot)$ means selecting the $k$ lagest elements in each row of the weighted semantic relation matrix $\mathcal{M}^{zn}$. 
	Typically, we set $k$ to 3 on the miniImagenet dataset. 
	Since the small diversity within the different classes, we set $k$ to 1 on three fine-grained datasets for our MATANet to capture the most discriminative features and avoid introducing noise.
	Under the N-way K-shot few-shot learning setting, we can get semantic similarity vectors $\mathcal{P}^z\in\mathbb{R}^N$.

		
		
	
	\subsection{Classification with an adaptive fusion strategy}
	
	Since the five relation scores have been calculated, we need to design a fusion module to integrate them. In order to solve this problem, we adopt a learnable five-dimensional $w=[w_1, w_2, w_3, w_4, w_5]$ vector to adaptively integrate these five parts. Specifically, the final fusion similarity between a query $\mathcal{A}^q_\mathcal{Q}$ and support set $\mathcal{S}$ can be defined as follows
	\begin{equation}
		\mathcal{P}^q = w_1 \cdot \mathcal{P}^{1}\ + w_2 \cdot 	\mathcal{P}^{2}+w_3 \cdot \mathcal{P}^{3}\ \\
		+ w_4 \cdot 	\mathcal{P}^{4} +w_5 \cdot 	\mathcal{P}^{5} 
	\end{equation}
	Under the 5-way 1-shot few-shot learning setting, input a query image, we will get  five similarity vectors $\mathcal{P}^{z}, z\in\left\{1,2,3,4,5\right\}$. We first balance the size of these five vectors by a Batch Normalization layer. Then, we concatenate these five vectors and use a 1D convolution layer with the kernel size of $1\times1$ and the dilation value of 5. Then we can get a weighted 5-dimensional similarity vector $\mathcal{P}^q$, we use it for final classification. 
		
	\section{Experiments}
	In this section, we perform extensive experiments to verify the advance and effectiveness of MATANet. 
	\subsection{Datasets}
	\textbf{\emph{mini}ImageNet.} As a small subset of ImageNet \cite{deng2009imagenet}, The dataset consists of 100 categories, each containing 600 images. We use common splits as in \cite{finn2017model}, which devides the dataset into training, validation and test dataset with 64/16/20 classes respectively.
	
	We also conduct experiments on three fine-grained image recognition datasets. 	
	
	\textbf{CUB Birds} \cite{wah2011caltech} is composed of 11, 788 images of 200 birds species. 
	\textbf{Stanford Dogs} \cite{khosla2011novel} is contains 120 categories of dogs and 20, 480 images. 
	\textbf{Stanford Cars} \cite{krause20133d} consists of 196 categories of cars with 16, 185 images. 	
	
	For fair comparisons, we strictly follow the splits used in \cite{li2019revisiting, li2019distribution} on Stanford Dogs and Stanford Cars, and follow the splits used in \cite{chen2019closer} on CUB Birds as Table 1 shows.
	\begin{table}[t]
		\centering
		\begin{tabular}{p{0.9cm}<{\centering}ccc}
			\toprule
			\textbf{Dataset} & \emph{\textbf{Stanford Dogs}}&\emph{\textbf{Stanford Cars}}&\emph{\textbf{CUB Birds}}
			\\
			\midrule
			$N_{all}$&120&196&200\\
			$N_{train}$&70&130&100\\
			$N_{val}$&20&17&50\\
			$N_{test}$&30&49&50\\
			\bottomrule
		\end{tabular}
		\caption{The splits of three fine-grained datasets. $N_{all}$ is the number of all classes. $N_{train}$, $N_{val}$ and $N_{test}$ indicate the number of classes in training set, validation set and test set.}
	\end{table}
	
	\begin{table*}[t]
		
		\centering
		\begin{tabular}{cccccc}
			\toprule
			\textbf{Model} &\textbf{Venue} &\textbf{Backbone} & \textbf{Type}  & \textbf{5-way 1-shot} & \textbf{5-way 5-shot}\\
			\midrule
			Meta LSTM$^*$ \cite{ravi2016optimization} &ICLR'17&  Conv-32F  &Meta& 43.44$\pm$\footnotesize{0.77} & 60.60$\pm$\footnotesize{0.71} \\
			MAML$^*$ \cite{finn2017model} & ICML'17 &Conv-32F  &Meta& 48.70$\pm$\footnotesize{1.84} & 63.11$\pm$\footnotesize{0.92} \\
			TAML-Entropy$^*$ \cite{jamal2019task} &CVPR'19 &  Conv-32F  &Meta& 49.33$\pm$\footnotesize{1.80}& 66.05$\pm$\footnotesize{0.85}\\
			MAML+L2F$^*$ \cite{baik2020learning} &CVPR'20 & Conv-32F  &Meta& 52.10$\pm$\footnotesize{0.49}& 69.38$\pm$\footnotesize{0.46}\\
			WarpGrad$^*$ \cite{flennerhag2019meta} &ICLR'20 &  Conv-32F  &Meta& \textbf{52.30}$\pm$\textbf{\footnotesize{0.80}}& 68.40$\pm$\footnotesize{0.60} \\
			
			\midrule
			Matching Nets$^*$ \cite{vinyals2016matching} & NeurIPS'16&Conv-64F & Metric  & 43.56$\pm$\footnotesize{0.84} &55.31$\pm$\footnotesize{0.73}  \\
			Prototypical Nets$^*$ \cite{snell2017prototypical} &NeurIPS'17& Conv-64F & Metric &49.42$\pm$\footnotesize{0.78} & 68.20$\pm$\footnotesize{0.66} \\
			Relation Nets$^*$ \cite{sung2018learning} & CVPR'18&Conv-64F & Metric &50.44$\pm$\footnotesize{0.82}&65.32$\pm$\footnotesize{0.70}\\
			GNN$^*$ \cite{garcia2017few} & CVPR'18&Conv-256F & Metric & 50.33$\pm$\footnotesize{0.36}& 66.41$\pm$\footnotesize{0.63} \\
			IMP$^*$ \cite{allen2019infinite}& ICML'19&Conv-64F & Metric &49.60$\pm$\footnotesize{0.80} & 68.10$\pm$\footnotesize{0.80} \\
			CovaMNet$^*$ \cite{li2019distribution} & AAAI'19&Conv-64F & Metric & 51.19$\pm$\footnotesize{0.76} &67.65$\pm$\footnotesize{0.63}\\\
			DN4$^*$ \cite{li2019revisiting} & CVPR'19&Conv-64F & Metric & 51.24$\pm$\footnotesize{0.74} &\textbf{71.02}$\pm$\textbf{\footnotesize{0.64}} \\
			SAML$^*$\cite{hao2019collect} & ICCV'19&Conv-64F & Metric & 52.22$\pm$\footnotesize{0.00}& 66.49$\pm$\footnotesize{0.00} \\
			DSN$^*$\cite{simon2020adaptive} & CVPR'20&Conv-64F & Metric & 51.78$\pm$\footnotesize{0.96}& 68.99$\pm$\footnotesize{0.69} \\
			
			\midrule
			\textbf{MATANet} &Ours& Conv-64F & Metric &\textbf{53.63}$\pm$\textbf{\footnotesize{0.83}}  & \textbf{72.67}$\pm$\textbf{\footnotesize{0.76}} \\
			\bottomrule
		\end{tabular}
		\caption{Comparison with other state-of-the-art methods with $95\%$ confidence intervals on mini-ImageNet. The third column shows which kind of embedding is employed. The fourth column shows which type of the method belongs to, i.e, meta-learning based, metric-learning based, and other kinds of methods. $^*$ Results reported by the original work. (Top two performances are in bold) }
	\end{table*}

	\subsection{Network architecture}
	It is a well-known fact that using deeper networks to extract features or using pre-trained models can achieve higher accuracy. We follow the basic feature extractor module which is adopted in previous works, to make a fair comparison with other works. The feature extractor module $\mathcal{F}_{\theta}$ consists of 4 convolutional blocks. Specifically, each convolutional block consists of a convolutional layer (with $3\times3$ convolution and 64 filters), a batch normalization layer, and a leaky ReLU non-linearity. Besides, we add a $2\times2$ max-pooling layer to the first two convolution blocks. The reason for using only two max-pooling layers is we can get more LRs to capture the semantic relation between them. For example, in a 5-way 1-shot few-shot learning task, if we use four maxpooling layers, we can only get 25 LRs for an $84\times84$ input image. In contrast, if we only use two max-pooling layers, we will get 441 LRs, which will be helpful for us to find local semantic relations.
	
	\subsection{Implementation Details}
	Our experiments are conducted under the N-way K-shot setting on four benchmarks. All the images in four benchmarks are resized to $84\times84$. During the training stage, we randomly construct 250, 000 episodes to train our MATANet for the  \emph{mini}ImageNet and Stanford Cars, and 300,000 for the other datasets by episodic training mechanism. In each episode, we select 15 or 10 query images from each class for the 1-shot or 5-shot setting, respectively, i.e., in a 5-way 1-shot task, we have 75 query images and 5 support images. We adopt the Adam algorithm \cite{kingma2014adam} with a cross-entropy (CE) loss to train the network. Also, the initial learning rate is set to 0.001 and reduce it by half of every 50,000 episodes. During the test stage, 600 episodes are constructed from the test set,  and this test process will be repeated five times. Then the mean accuracy and $95\%$ confidence intervals will be reported simultaneously.

	\subsection{Baselines}
	To evaluate the effectiveness of our MATANet on the \emph{mini}ImageNet dataset, we make comparisons with state-of-the art methods. Since our method is a metric-learning based method, we mainly compare our MATANet with methods in this branch, including Matching Nets \cite{vinyals2016matching}, Prototypical Nets \cite{snell2017prototypical}, Relation Nets \cite{sung2018learning}, GNN \cite{garcia2017few}, IMP \cite{allen2019infinite}, CovaMNet \cite{li2019distribution}, DN4 \cite{li2019revisiting}, SAML\cite{hao2019collect} and DSN \cite{simon2020adaptive} . We also pick five meta-learning models Meta LSTM \cite{ravi2016optimization}, MAML \cite{finn2017model}, TAML-Entropy \cite{jamal2019task}, MAML+L2F \cite{baik2020learning} and WarpGrad \cite{flennerhag2019meta} for reference. 
	
	We compare seven few-shot leraning methods on fine-grained datasets, Matching Nets \cite{vinyals2016matching}, Prototypical Nets \cite{snell2017prototypical}, MAML \cite{finn2017model}, Relation Nets  \cite{sung2018learning}, CovaMNet \cite{li2019distribution}, DN4 \cite{li2019revisiting}, and $\rm {PABN_{+cpt}}$/$\rm {LRPABN_{+cpt}}$ \cite{huang2020low}.

	\begin{table*}
		\centering
		\begin{tabular}{c  p{1.8cm}<{\centering}  p{1.8cm}<{\centering}  p{1.8cm}<{\centering}  p{1.8cm}<{\centering}  p{1.8 cm}<{\centering}  p{1.8cm}<{\centering} }
			\toprule
			\multirow{3}{*}{\textbf{Model}} & \multicolumn{6}{c}{\textbf{5-Way Accuracy($\%$)}}
			\\
			\cmidrule{2-7}
			&\multicolumn{2}{c}{\emph{\textbf{Stanford Dogs}}} &\multicolumn{2}{c}{\emph{\textbf{Stanford Cars}}} 
			&\multicolumn{2}{c}{\emph{\textbf{CUB Birds}}} \\
			& 1-shot& 5-shot & 1-shot & 5-shot & 1-shot & 5shot\\
			\midrule
			Matching Nets$^\dagger$ \cite{vinyals2016matching} & 35.80$\pm$\footnotesize{0.99} & 47.50$\pm$\footnotesize{1.03}  & 34.80$\pm$\footnotesize{0.98} & 44.70$\pm$\footnotesize{1.03}  & 61.16$\pm$\footnotesize{0.89} & 72.86$\pm$\footnotesize{0.70} \\
			Prototypical Nets$^\dagger$ \cite{snell2017prototypical}& 37.59$\pm$\footnotesize{1.00} & 48.19$\pm$\footnotesize{1.03}  & 40.90$\pm$\footnotesize{1.01} & 52.93$\pm$\footnotesize{1.03} & 51.31$\pm$\footnotesize{0.91} & 70.77$\pm$\footnotesize{0.69}  \\
			MAML$^\ddagger$ \cite{finn2017model} & 44.81$\pm$\footnotesize{0.34} & 58.68$\pm$\footnotesize{0.31} &47.22$\pm$\footnotesize{0.39} & 61.21$\pm$\footnotesize{0.28} & 55.92$\pm$\footnotesize{0.95}  &  72.09$\pm$\footnotesize{0.76} \\
			Relation Nets$^\ddagger$ \cite{sung2018learning} & 43.33$\pm$\footnotesize{0.42} & 55.23$\pm$\footnotesize{0.41} & 47.67$\pm$\footnotesize{0.47}  & 60.59$\pm$\footnotesize{0.40} & 62.45$\pm$\footnotesize{0.98}  & 76.11$\pm$\footnotesize{0.69} \\
			CovaMNet$^*$  \cite{li2019distribution} & \textbf{49.10}$\pm$\textbf{\footnotesize{0.76}}  & 63.04$\pm$\footnotesize{0.65} & 56.65$\pm$\footnotesize{0.86} & 71.33$\pm$\footnotesize{0.62}  & 60.58$\pm$\footnotesize{0.69} & 74.24$\pm$\footnotesize{0.68} \\
			DN4$^*$  \cite{li2019revisiting} & 45.41$\pm$\footnotesize{0.76} & \textbf{63.51}$\pm$\textbf{\footnotesize{0.62}}  & 59.84$\pm$\footnotesize{0.80} & \textbf{88.65}$\pm$\textbf{\footnotesize{0.44}}  & 52.79$\pm$\footnotesize{0.86} & \textbf{81.45}$\pm$\textbf{\footnotesize{0.70}}  \\
			\midrule
			$\rm {PABN_{+cpt}}^*$ \cite{huang2020low} & 45.65$\pm$\footnotesize{0.71} & 61.24$\pm$\footnotesize{0.62} & 54.44$\pm$\footnotesize{0.71}  & 67.36$\pm$\footnotesize{0.61}  & 63.56$\pm$\footnotesize{0.79} & 75.35$\pm$\footnotesize{0.58} \\
			$\rm {LRPABN_{+cpt}}^*$ \cite{huang2020low} & 45.72$\pm$\footnotesize{0.75} & 60.94$\pm$\footnotesize{0.66} & \textbf{60.28}$\pm$\textbf{\footnotesize{0.76}} & 73.29$\pm$\footnotesize{0.58} & \textbf{63.63}$\pm$\textbf{\footnotesize{0.77}} & 76.06$\pm$\footnotesize{0.58} \\		
			\midrule 
			\textbf{MATANet(Ours)} & \textbf{55.63}$\pm$\textbf{\footnotesize{0.88}} & \textbf{70.29}$\pm$\textbf{\footnotesize{0.62}}   & \textbf{73.15}$\pm$\textbf{\footnotesize{0.88}}  & \textbf{91.89}$\pm$\textbf{\footnotesize{0.45}} &  \textbf{67.33}$\pm$\textbf{\footnotesize{0.84}}  & \textbf{83.92} $\pm$\textbf{\footnotesize{0.63}} \\
			\bottomrule
		\end{tabular}
		\caption{Experimental results compared with other methods on three fine-grained datasets. For Stanford Dog and Stanford Car, $^*$ results reported by the original work, $^\dagger$ results reported by \cite{li2019revisiting}, $^\ddagger$ results re-implemented in the same setting for a fair comparison. For CUB Birds, we adopt the results for first four methods from \cite{chen2019closer}, re-implement CovaMNet \cite{li2019distribution} and DN4 \cite{li2019revisiting}, and adopt the results reported by the original work for $\rm {PABN_{+cpt}}$/$\rm {LRPABN_{+cpt}}$ \cite{huang2020low}. (Top two performances are in bold)}
	\end{table*}

	\begin{table*}[t]
		\centering
		\begin{tabular}{cccccc}
			\toprule
			\multirow{2}{*}{\textbf{Model}}&\multirow{2}{*}{\textbf{Venue}} &\multirow{2}{*}{\textbf{Backbone}}&\multirow{2}{*}{\textbf{Type}}& \multicolumn{2}{c}{\textbf{5-Way Accuracy($\%$)}}
			\\
			\cmidrule{5-6} 
			& &&&1-shot& 5-shot \\
			\midrule
			Dynamic-Net$^{*}$ \cite{gidaris2018dynamic}&CVPR'18&ResNet12 &Meta&55.45$\pm$\footnotesize{0.89} &70.13$\pm$\footnotesize{0.68} 
			\\
			SNAIL$^{*}$ \cite{mishra2017simple}&ICLR'18&ResNet12 &Meta&55.71$\pm$\footnotesize{0.99} &68.88$\pm$\footnotesize{0.92} 
			\\
			TPN$^{*}$ \cite{liu2018learning}&ICLR‘19&ResNet12 &Others&59.46$\pm$\footnotesize{0.00} &75.64$\pm$\footnotesize{0.00} 
			\\	
			MAML+L2F$^*$ \cite{baik2020learning} &CVPR‘20&ResNet12 &Meta&57.48$\pm$\footnotesize{0.49} &74.68$\pm$\footnotesize{0.43} 
			\\	
			\midrule
			\textbf{MATANet} &Ours&ResNet12 &Metric& \textbf{60.13}$\pm$\textbf{\footnotesize{0.81}}& \textbf{75.42}$\pm$\textbf{\footnotesize{0.72}}
			\\
			\midrule
			Qiao$^{*}$ \cite{qiao2018few}&CVPR'18&WRN-28-10&Meta&59.60$\pm$\footnotesize{0.41} &73.74$\pm$\footnotesize{0.19} \\

			LEO-trainal$^{*}$ \cite{rusu2018meta}&ICLR'19&WRN-28-10&Meta&61.76$\pm$\footnotesize{0.08} &77.59$\pm$\footnotesize{0.12} 
			\\
			Fine-tuning$^{*}$ \cite{dhillon2019baseline}&ICLR'20&WRN-28-10&Others&57.73$\pm$\footnotesize{0.62} &78.17$\pm$\footnotesize{0.49} 
			\\
			LEO+L2F$^*$ \cite{baik2020learning} &CVPR‘20&WRN-28-10 &Meta&62.12$\pm$\footnotesize{0.13} &78.13$\pm$\footnotesize{0.15} 
			\\	
			\midrule
			\textbf{MATANet}&Ours&WRN-28-10&Metric& \textbf{62.43}$\pm$\textbf{\footnotesize{0.78}}& \textbf{79.02}$\pm$\textbf{\footnotesize{0.72}}
			\\
			\bottomrule
		\end{tabular}
		\caption{
			Comparison with other state-of-the-art methods that use deeper backbones with $95\%$ confidence intervals on mini-ImageNet. The third column shows which kind of embedding is employed. The fourth column shows which type of method belongs to. $^*$ Results reported by the original work.}
	\end{table*}

	\subsection{Comparisons with the SOTA Methods}
	Our method is compared with several state-of-the-art methods under 5-way 1-shot and 5-way 5-shot few-shot learning settings.
	
	\textbf{Results on \emph{mini}ImageNet.} The experimental results on \emph{mini}ImageNet are reported in Table 2.  It can be observed that our method significantly outperforms other methods under both 5-way 1-shot and 5-shot settings. 
	Especially, we are 2.5\% better than the second best method \cite{flennerhag2019meta} under the 5-way 1-shot setting, with an accuracy rate of 53.63\%.  Similarly, we achieve 72.67\% under the 5-way 5-shot setting, with an improvement of 2.3\% from the second best method \cite{li2019revisiting}. Note that, our model gains 4.7\% and 2.3\% improvements over the most relevant work \cite{li2019revisiting} on 1-shot and 5-shot, respectively, which proposes an image-to-class mechanism to find the relation at class-level. This improvement verifies the effectiveness of our model, which can adaptively select the most discriminative local features at multiple scales in a certain task.
	
	\textbf{Results on fine-grained datasets.} From Table 3, it can be observed that the proposed MATANet outperforms all other state-of-the-art methods under both 5-way 1-shot and 5-way 5-shot few-shot learning settings. 
	Especially for the 5-way 1-shot task, our method achieves 13.3\%, 21.6\%, and 5.8\% gains over the second best on Stanford Dogs, Stanford Cars, and CUB Birds, respectively. For the 5-way 5-shot task, our method achieves 10.7\%, 3.7\%, and 3.0\% gains over the second best on three datasets.
	The reason why we can achieve these state-of-the-art performances is that MATANet can adaptively select the task-relevant LRs at multiple scales for classification. 

		\begin{figure}[t]
		\centering
		\includegraphics[height=5cm,width=7cm]{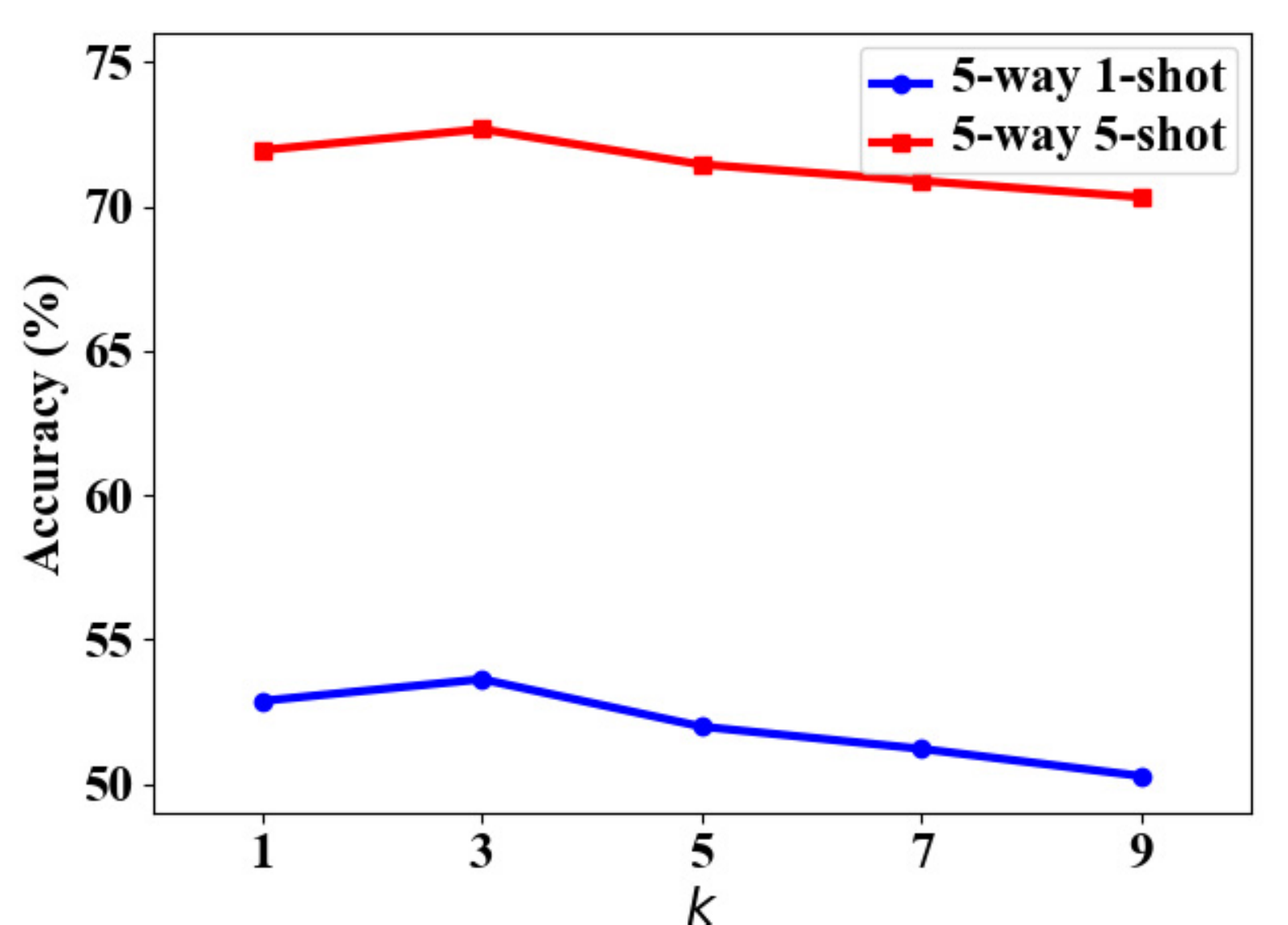}
		\caption{Experimental results of MATANet using different superparameter $k$ on \emph{mini}ImageNet. }
		\label{influence k}
	\end{figure}
	\begin{figure*}[t]
		\centering
		\includegraphics[height=8.8cm,width=15.5cm]{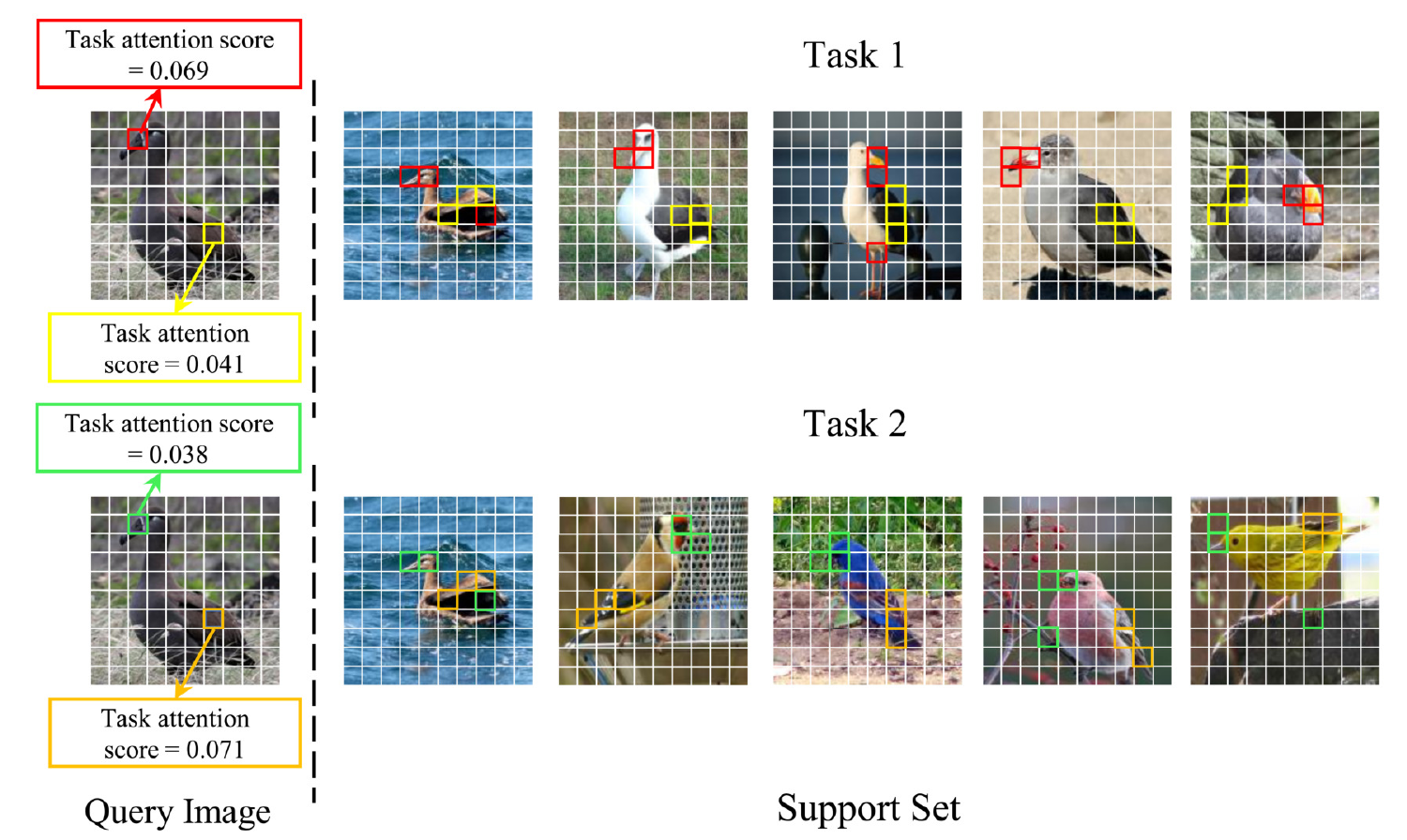}
		\caption{Visualization of the selected LRs. In a 5-way 1-shot task, for each LR of the query image, our method can find the $k$ (i.e., $k$=3) most similar LRs of all support LRs for a certain class, and weight them according to the importance of the query LR in the current task. }
		\label{k}
	\end{figure*}

\subsection{Discussion}
	\textbf{Influence of superparameter $\bm{k}$.}
	The results of the empirical study about using different $k$ in MATANet.
	In the similarity-to-class module, for each LR of the query image, we need to search out the $k$ most similar LRs of all support LRs in each class. Next, we adaptively integrate five relation scores that be calculated at different scales for final prediction. How to choose a suitable $k$ is particularly significant. For this purpose, we conduct a contrast experiment on \emph{mini}ImageNet dataset under both 5-way 1-shot and 5-way 5-shot settings by varying the value of $k\in \left\{1,3,5,7,9\right\}$. As shown in Figure 4, the value of $k$ has a moderate influence on classification performance, so we should choose a specific $k$ for each task.

\begin{table}[t]
	\centering
	\begin{tabular}{ccc}
		\toprule
		\multirow{2}{*}{\textbf{Metric functions}} & \multicolumn{2}{c}{\textbf{5-Way Accuracy($\%$)}}
		\\
		\cmidrule{2-3} 
		& 1-shot& 5-shot \\
		\midrule
		$e^{-d(a,b)}$&52.82$\pm$\footnotesize{0.83} &71.85$\pm$\footnotesize{0.74} 
		\\
		$\frac{1}{1+d(a,b)}$ &52.11$\pm$\footnotesize{0.85} &71.25$\pm$\footnotesize{0.76} 
		\\
		Tanimito Index &53.25$\pm$\footnotesize{0.84} &72.11$\pm$\footnotesize{0.73} 
		\\
		Cosine Similarity&53.63$\pm$\footnotesize{0.83} &	72.67$\pm$\footnotesize{0.76} 	
		\\
		\bottomrule
	\end{tabular}
	\caption{Experimental results of MATANet using different metric functions on \emph{mini}ImageNet, $d(a,b)$:the euclidean distance between vector $a$ and $b$. 
	}
\end{table}

	\textbf{Influence of backbone networks.} Besides the simple Conv-64F, we also use other deeper feature extractors to evaluate our model, i.e., ResNet12 and WRN-28-10. We compared other state-of-the-art methods that using these deeper feature extractors, including Dynamic-Net \cite{gidaris2018dynamic}, SNAIL \cite{mishra2017simple}, TPN \cite{liu2018learning}, MAML+L2F \cite{baik2020learning}, Qiao \cite{qiao2018few}, LEO \cite{rusu2018meta}, Fine-tuning \cite{dhillon2019baseline} and LEO+L2F \cite{baik2020learning}. When using deeper feature extractors, the accacy of MATANet reaches 60.13\% and 62.43\% for the 5-way 1-shot task, 75.42\% and 79.13\% for the 5-shot task, by using ResNet12 and WRN-28-10, repectively. Moreover, when using same deeper feature exactor, our MATANet outperforms all other methods under both 5-way 1-shot and 5-way 5-shot few-shot learning settings (see Table 4).
	
	\textbf{Influence of metric functions.} The results on the different metric functions using in Adaptive Task Attention Module are reported in Table 5, and the Tanimito Index is represented as $T(a,b) = \frac{a \cdot b}{\left\|a\right\| \cdot \left\|b\right\|-a\cdot b }$. To measure the semantic relations between feature descriptors, a suitable metric functions is a key factors. It can be seen in Table 5, the best metric function is Cosine Similarity.

	\textbf{Ablation study.} To further verify the effectiveness of the multi-scale feature generator, adaptive task-attention module, and similarity-to-class module, we perform an ablation study on \emph{mini}ImageNet. We remove $g_{\phi}$, $\mathcal{F}_{\varphi}$ and $\mathcal{F}_{\omega}$  from the MATANet respectively to confirm that each part of the model is indispensable. We remove $g_{\phi}$, $\mathcal{F}_{\varphi}$ and $\mathcal{F}_{\omega}$ simultaneously as the baseline method. As seen in Table 6, the main improvement comes from the adaptive task-attention module $\mathcal{F}_{\varphi}$. If we remove $\mathcal{F}_{\varphi}$, the performance will be reduced by $3.7\%$, $4.8\%$ on 1-shot, 5-shot tasks, respectively. This empirical study proves that the discriminative $\mathcal{F}_{\varphi}$ gives a performance boost and results in more discriminative features for classification. Similarly, if we remove $g_{\phi}$, the performance will be reduced by $1.3\%$, $0.9\%$ on 1-shot, 5-shot tasks, respectively. Moreover, if we remove $\mathcal{F}_{\omega}$, the performance will be reduced by $1.1\%$, $0.8\%$ on 1-shot, 5-shot tasks, respectively. 

\begin{table}[t]
	\centering
	\begin{tabular}{ccc}
		\toprule
		\multirow{2}{*}{\textbf{Model}} & \multicolumn{2}{c}{\textbf{5-Way Accuracy($\%$)}}
		\\
		\cmidrule{2-3} 
		& 1-shot& 5-shot \\
		\midrule
		baseline&51.12$\pm$\footnotesize{0.73} &67.35$\pm$\footnotesize{0.63} 
		\\	
		w/o $g_{\phi}$&52.95$\pm$\footnotesize{0.77} &72.02$\pm$\footnotesize{0.69} \\
		w/o $\mathcal{F}_{\varphi}$&51.66$\pm$\footnotesize{0.77} &69.12$\pm$\footnotesize{0.77} 	\\
		w/o $\mathcal{F}_{\omega}$&53.02$\pm$\footnotesize{0.79} &72.07$\pm$\footnotesize{0.73} 	\\
		\midrule
		\textbf{MATANet(ours)} &\textbf{53.63}$\pm$\textbf{\footnotesize{0.83}} &\textbf{72.67}$\pm$\textbf{\footnotesize{0.76}}
		\\	
		\bottomrule
	\end{tabular}
	\caption{The ablation study on \emph{mini}ImageNet for the proposed MATANet.}
\end{table}	
	
	\textbf{Visualization of the selected LRs.} As shown in Figure 5, for the LRs in red, yellow, green and orange boxes in the query image, we visualized the $k$ (i.e., $k$=3) most discriminative LRs selected by MATANet. In \cite{li2019revisiting}, they will equally use the selected LRs for the final classification. 
	However, in our method, the LRs corresponding to these boxes are treated differently. In task 1, the task attention score corresponding to the red box and yellow box is 0.069 and 0.041, respectively, so the LRs corresponding to the red box will play a more important role in the final classification. Similarly,  in task 2, the LRs corresponding to the orange box will play a more important role in the final classification. This is because the beak is obviously more discriminative than the wing in task 1. While in Task 2 the wing is significantly more discriminative than the beak, which verifies that our method can automatically select the most discriminative LRs in the current task. Moreover, it can be seen from Figure 5 that the scales of the dominant objects in different images are different, which may affect the performance of the model. This once again verified the necessity of our multi-scale feature generator.
		
	\section{Conclusion}
	In this paper, we revisit the local representation based metric-learning and propose a novel Multi-scale Adaptive Task Attention Network (MATANet) for few-shot learning, aiming to learn more discriminative task-relevant local representations at different scales by generating multiple features at different scale and looking at the context of the entire task. 
	By taking a view of the entire task, our method is able to adaptively select the most discriminative local representations in the current task at different scales. 
	Extensive experiments on four benchmark datasets demonstrate the effectiveness and advantages of the proposed MATANet. 

	{\small
		\bibliographystyle{ieee_fullname}
		\bibliography{MATANet}

\begin{thebibliography}{10}\itemsep=-1pt

\bibitem{allen2019infinite}
Kelsey~R Allen, Evan Shelhamer, Hanul Shin, and Joshua~B Tenenbaum.
\newblock Infinite mixture prototypes for few-shot learning.
\newblock {\em arXiv preprint arXiv:1902.04552}, 2019.

\bibitem{baik2020learning}
Sungyong Baik, Seokil Hong, and Kyoung~Mu Lee.
\newblock Learning to forget for meta-learning.
\newblock In {\em Proceedings of the IEEE/CVF Conference on Computer Vision and
  Pattern Recognition}, pages 2379--2387, 2020.

\bibitem{chen2019closer}
Wei-Yu Chen, Yen-Cheng Liu, Zsolt Kira, Yu-Chiang~Frank Wang, and Jia-Bin
  Huang.
\newblock A closer look at few-shot classification.
\newblock {\em arXiv preprint arXiv:1904.04232}, 2019.

\bibitem{deng2009imagenet}
Jia Deng, Wei Dong, Richard Socher, Li-Jia Li, Kai Li, and Li Fei-Fei.
\newblock Imagenet: A large-scale hierarchical image database.
\newblock In {\em 2009 IEEE conference on computer vision and pattern
  recognition}, pages 248--255. Ieee, 2009.

\bibitem{dhillon2019baseline}
Guneet~S Dhillon, Pratik Chaudhari, Avinash Ravichandran, and Stefano Soatto.
\newblock A baseline for few-shot image classification.
\newblock {\em arXiv preprint arXiv:1909.02729}, 2019.

\bibitem{finn2017model}
Chelsea Finn, Pieter Abbeel, and Sergey Levine.
\newblock Model-agnostic meta-learning for fast adaptation of deep networks.
\newblock {\em arXiv preprint arXiv:1703.03400}, 2017.

\bibitem{flennerhag2019meta}
Sebastian Flennerhag, Andrei~A Rusu, Razvan Pascanu, Francesco Visin, Hujun
  Yin, and Raia Hadsell.
\newblock Meta-learning with warped gradient descent.
\newblock {\em arXiv preprint arXiv:1909.00025}, 2019.

\bibitem{garcia2017few}
Victor Garcia and Joan Bruna.
\newblock Few-shot learning with graph neural networks.
\newblock {\em arXiv preprint arXiv:1711.04043}, 2017.

\bibitem{gidaris2018dynamic}
Spyros Gidaris and Nikos Komodakis.
\newblock Dynamic few-shot visual learning without forgetting.
\newblock In {\em Proceedings of the IEEE Conference on Computer Vision and
  Pattern Recognition}, pages 4367--4375, 2018.

\bibitem{hao2019collect}
Fusheng Hao, Fengxiang He, Jun Cheng, Lei Wang, Jianzhong Cao, and Dacheng Tao.
\newblock Collect and select: Semantic alignment metric learning for few-shot
  learning.
\newblock In {\em Proceedings of the IEEE International Conference on Computer
  Vision}, pages 8460--8469, 2019.

\bibitem{he2016deep}
Kaiming He, Xiangyu Zhang, Shaoqing Ren, and Jian Sun.
\newblock Deep residual learning for image recognition.
\newblock In {\em Proceedings of the IEEE conference on computer vision and
  pattern recognition}, pages 770--778, 2016.

\bibitem{huang2020low}
Huaxi Huang, Junjie Zhang, Jian Zhang, Jingsong Xu, and Qiang Wu.
\newblock Low-rank pairwise alignment bilinear network for few-shot
  fine-grained image classification.
\newblock {\em IEEE Transactions on Multimedia}, 2020.

\bibitem{jamal2019task}
Muhammad~Abdullah Jamal and Guo-Jun Qi.
\newblock Task agnostic meta-learning for few-shot learning.
\newblock In {\em Proceedings of the IEEE Conference on Computer Vision and
  Pattern Recognition}, pages 11719--11727, 2019.

\bibitem{khosla2011novel}
Aditya Khosla, Nityananda Jayadevaprakash, Bangpeng Yao, and Fei-Fei Li.
\newblock Novel dataset for fine-grained image categorization: Stanford dogs.
\newblock In {\em Proc. CVPR Workshop on Fine-Grained Visual Categorization
  (FGVC)}, volume~2, 2011.

\bibitem{kingma2014adam}
Diederik~P Kingma and Jimmy Ba.
\newblock Adam: A method for stochastic optimization.
\newblock {\em arXiv preprint arXiv:1412.6980}, 2014.

\bibitem{koch2015siamese}
Gregory Koch, Richard Zemel, and Ruslan Salakhutdinov.
\newblock Siamese neural networks for one-shot image recognition.
\newblock In {\em ICML deep learning workshop}, volume~2. Lille, 2015.

\bibitem{krause20133d}
Jonathan Krause, Michael Stark, Jia Deng, and Li Fei-Fei.
\newblock 3d object representations for fine-grained categorization.
\newblock In {\em Proceedings of the IEEE international conference on computer
  vision workshops}, pages 554--561, 2013.

\bibitem{li2019revisiting}
Wenbin Li, Lei Wang, Jinglin Xu, Jing Huo, Yang Gao, and Jiebo Luo.
\newblock Revisiting local descriptor based image-to-class measure for few-shot
  learning.
\newblock In {\em Proceedings of the IEEE Conference on Computer Vision and
  Pattern Recognition}, pages 7260--7268, 2019.

\bibitem{li2019distribution}
Wenbin Li, Jinglin Xu, Jing Huo, Lei Wang, Yang Gao, and Jiebo Luo.
\newblock Distribution consistency based covariance metric networks for
  few-shot learning.
\newblock In {\em Proceedings of the AAAI Conference on Artificial
  Intelligence}, volume~33, pages 8642--8649, 2019.

\bibitem{liu2018learning}
Yanbin Liu, Juho Lee, Minseop Park, Saehoon Kim, Eunho Yang, Sung~Ju Hwang, and
  Yi Yang.
\newblock Learning to propagate labels: Transductive propagation network for
  few-shot learning.
\newblock {\em arXiv preprint arXiv:1805.10002}, 2018.

\bibitem{mishra2017simple}
Nikhil Mishra, Mostafa Rohaninejad, Xi Chen, and Pieter Abbeel.
\newblock A simple neural attentive meta-learner.
\newblock {\em arXiv preprint arXiv:1707.03141}, 2017.

\bibitem{qiao2018few}
Siyuan Qiao, Chenxi Liu, Wei Shen, and Alan~L Yuille.
\newblock Few-shot image recognition by predicting parameters from activations.
\newblock In {\em Proceedings of the IEEE Conference on Computer Vision and
  Pattern Recognition}, pages 7229--7238, 2018.

\bibitem{ravi2016optimization}
Sachin Ravi and Hugo Larochelle.
\newblock Optimization as a model for few-shot learning.
\newblock 2016.

\bibitem{rusu2018meta}
Andrei~A Rusu, Dushyant Rao, Jakub Sygnowski, Oriol Vinyals, Razvan Pascanu,
  Simon Osindero, and Raia Hadsell.
\newblock Meta-learning with latent embedding optimization.
\newblock {\em arXiv preprint arXiv:1807.05960}, 2018.

\bibitem{santoro2016meta}
Adam Santoro, Sergey Bartunov, Matthew Botvinick, Daan Wierstra, and Timothy
  Lillicrap.
\newblock Meta-learning with memory-augmented neural networks.
\newblock In {\em International conference on machine learning}, pages
  1842--1850, 2016.

\bibitem{simon2020adaptive}
Christian Simon, Piotr Koniusz, Richard Nock, and Mehrtash Harandi.
\newblock Adaptive subspaces for few-shot learning.
\newblock In {\em Proceedings of the IEEE/CVF Conference on Computer Vision and
  Pattern Recognition}, pages 4136--4145, 2020.

\bibitem{snell2017prototypical}
Jake Snell, Kevin Swersky, and Richard Zemel.
\newblock Prototypical networks for few-shot learning.
\newblock In {\em Advances in neural information processing systems}, pages
  4077--4087, 2017.

\bibitem{sun2019meta}
Qianru Sun, Yaoyao Liu, Tat-Seng Chua, and Bernt Schiele.
\newblock Meta-transfer learning for few-shot learning.
\newblock In {\em Proceedings of the IEEE conference on computer vision and
  pattern recognition}, pages 403--412, 2019.

\bibitem{sung2018learning}
Flood Sung, Yongxin Yang, Li Zhang, Tao Xiang, Philip~HS Torr, and Timothy~M
  Hospedales.
\newblock Learning to compare: Relation network for few-shot learning.
\newblock In {\em Proceedings of the IEEE Conference on Computer Vision and
  Pattern Recognition}, pages 1199--1208, 2018.

\bibitem{vinyals2016matching}
Oriol Vinyals, Charles Blundell, Timothy Lillicrap, Daan Wierstra, et~al.
\newblock Matching networks for one shot learning.
\newblock In {\em Advances in neural information processing systems}, pages
  3630--3638, 2016.

\bibitem{wah2011caltech}
Catherine Wah, Steve Branson, Peter Welinder, Pietro Perona, and Serge
  Belongie.
\newblock The caltech-ucsd birds-200-2011 dataset.
\newblock 2011.

\bibitem{zagoruyko2016wide}
Sergey Zagoruyko and Nikos Komodakis.
\newblock Wide residual networks.
\newblock {\em arXiv preprint arXiv:1605.07146}, 2016.

\end{thebibliography}
	}
	\clearpage
\section*{Supplementary Material }

\noindent
\textbf{More Visualization Results}

In our article, we visualized the $k$ (i.e., $k$=3) most discriminative LRs selected by MATANet. This is achieved by outputting their index and corresponding task attention values. 

We provide some classification examples in Figure 6, compare with the most relevant work DN4. By sending a fixed testing batch through the model, which consists of one support sample and five query samples for each of five classes, the prediction of MATANet only contains 7 mislabels in the entire 25 queries, while the prediction of DN4 has 10 wrong labels. That validates the effectiveness of the MATANet. Our model is able to adaptively select the most discriminative local representations in the current task at different scales. We also find that in some classes like Black-footed Akbatross and Rhinoceros Auklet, the high intra-variance and low inter-variance confuse all the models.

\begin{figure}[h]
	\centering          
	\subfigure[by DN4]{       \begin{minipage}{8.5cm}      \centering      
			\includegraphics[height = 5.8cm, width=8cm]{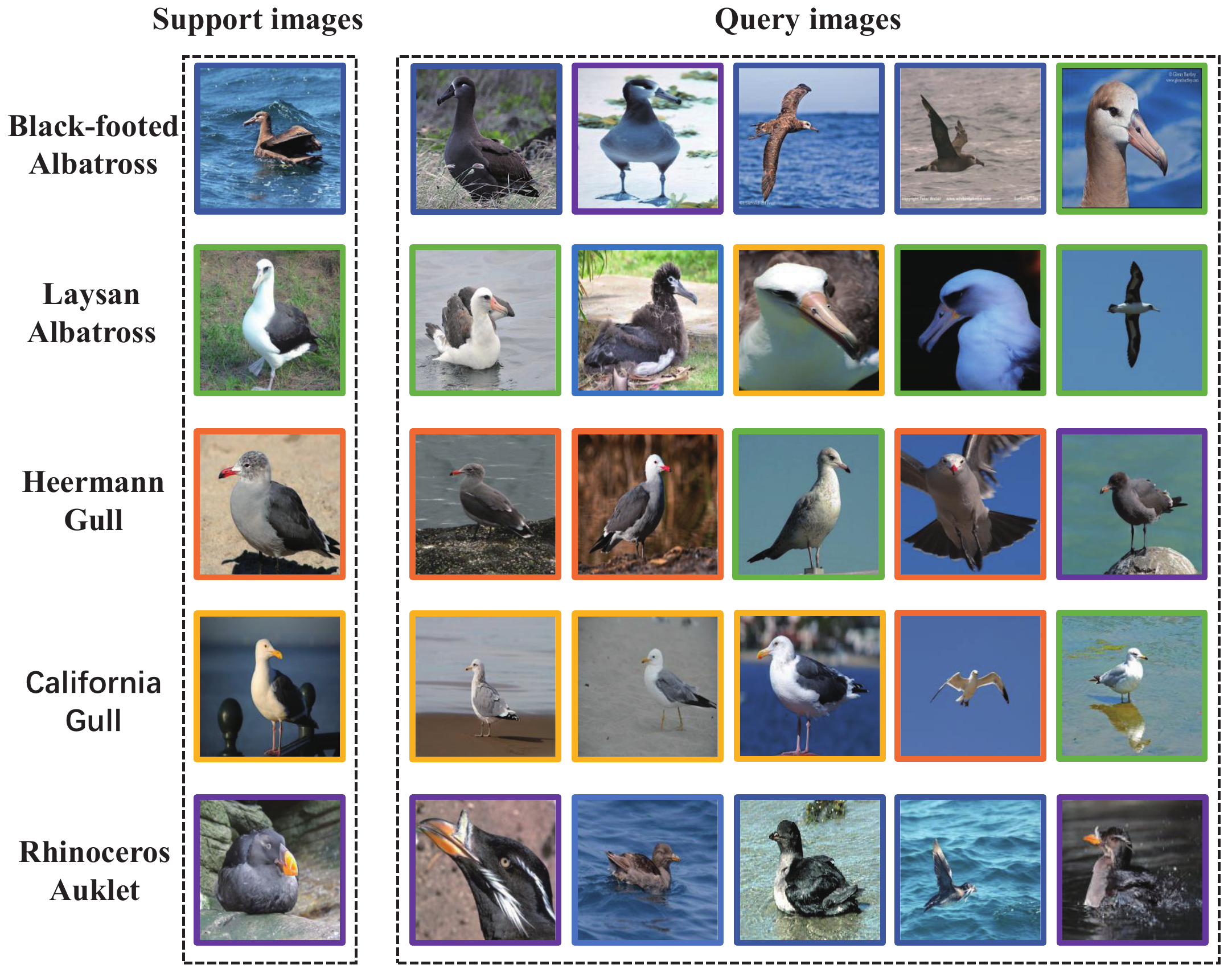}      
	\end{minipage}      }      
	\subfigure[by MATANet]{       
		\begin{minipage}{8.5cm}      
			\centering      
			\includegraphics[height = 5.8cm, width = 8 cm]{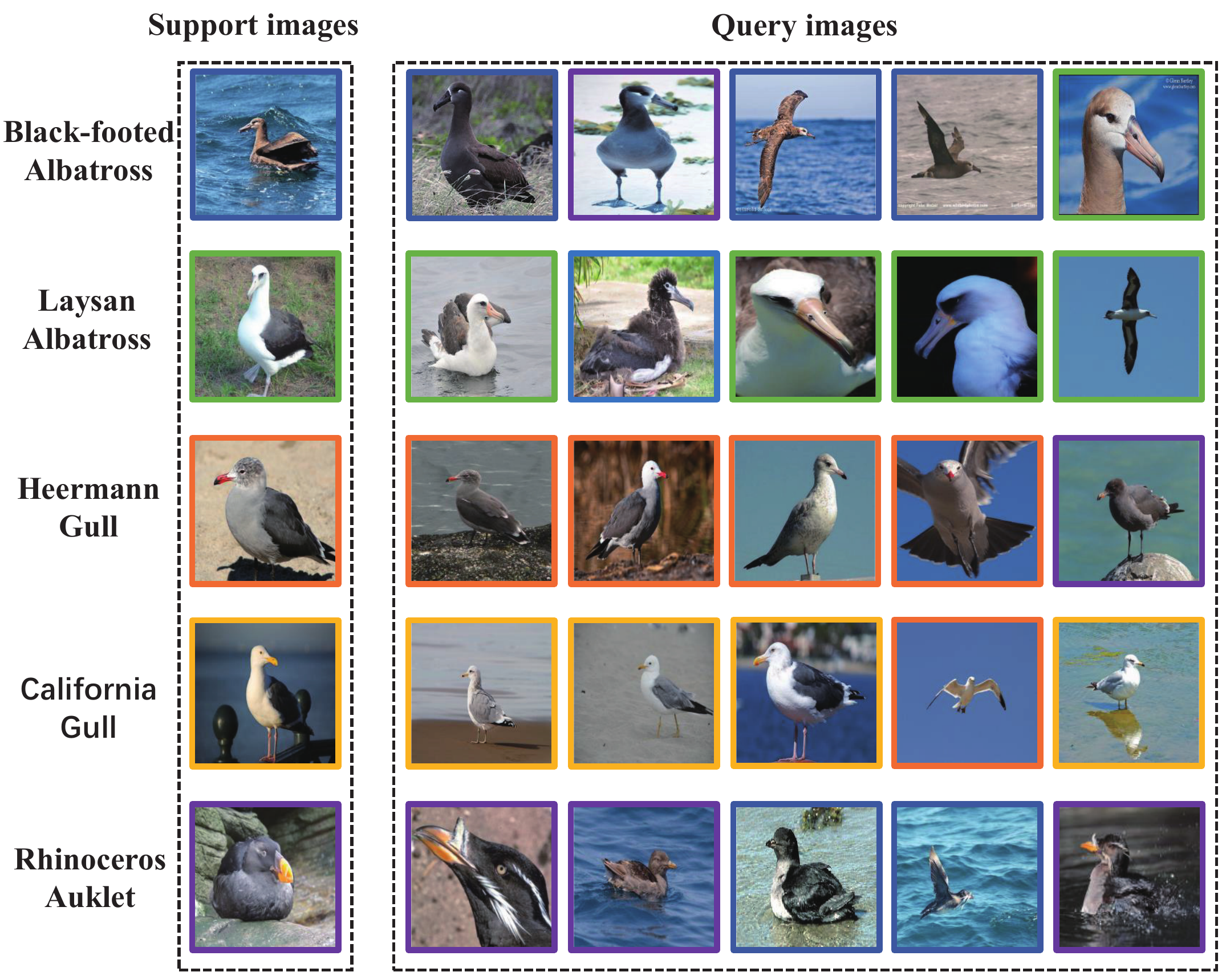}       
	\end{minipage}      }      \caption{Some visual classification results of comparing methods over CUB Birds dataset. Both DN4 and MATANet use the same data batch under the 5-way 1-shot setting, and for each class, we randomly select five query images as the testing data. We adopt five colors to label the support classes separately. As to the query images, we label the images with the color corresponding to the class label predicted by different models.}      \label{fig2}  
\end{figure}

\noindent
\textbf{The Number of Trainable Parameters}

We compare the number of trainable parameters to verify the efficiency of the proposed MATANet, as Table 7 shows. Since no other trainable parameters are introduced except for the embedding module $\mathcal{F}_{\theta}$ Prototypical Nets and DN4 become the most light-weight models. GNN adopts a larger embedding module (i.e., the filter number is 256), which draws a great contribution from the number of parameters. Relation Network and CovaMNet adopt additional architectures to boost the result, which also introduces a huge number of trainable parameters. However, the proposed MATANet only introduces a small number of the trainable parameters, while achieves a better result than the methods above.
\begin{table}[h]
	\centering
	\begin{tabular}{ccc}
		\toprule
		\textbf{Model}
		& \textbf{Params}& \textbf{Accuracy(\%)} \\
		\midrule
		\textbf{Prototypical Net}& \textbf{0.113M} &49.42$\pm$\footnotesize{0.78}
		\\
		\textbf{Relation Net} &0.229M &50.44$\pm$\footnotesize{0.82}
		\\
		\textbf{GNN}&1.619M &50.33$\pm$\footnotesize{0.36}
		\\
		\textbf{DN4}&\textbf{0.113M} &51.24$\pm$\footnotesize{0.74}
		\\
		\midrule
		\textbf{MATANet(Ours)}&0.314M& \textbf{53.63}$\pm$\textbf{\footnotesize{0.83}}\\
		\bottomrule
	\end{tabular}
	\caption{The number of trainable parameters in different models and the corresponding classification accuracies on \emph{mini}ImageNet under 5-way 1-shot setting.
	}
\end{table}

\noindent
\textbf{Tranining procedure}

The training procedure of the proposed MATANet is shown in Algorithm 1.

\begin{algorithm}[h]
	\caption{Tranining procedure}
	\label{Algo}	
	
	\hangafter=1
	\setlength{\hangindent}{2em}
	
	\textbf{Input:} Eposidic task $\mathcal{T}= \left\{\mathcal{A_S}, \mathcal{A_Q}\right\}$, superparameter $k$.
	\begin{algorithmic}[h]
		\While{no converge}
		\For{$\mathcal{A}^q_\mathcal{Q}$ in $\mathcal{A}_\mathcal{Q}$}
		\State $\mathcal{L}^{qz}\leftarrow\mathcal{G}_{\phi}(\mathcal{F}_{\theta}(\mathcal{A_Q}))$
		\State
		$\mathcal{L}^{\mathcal{S}z}\leftarrow\mathcal{G}_{\phi}(\mathcal{F}_{\theta}(\mathcal{S}))$
		\State Obtain semantic relation matrix $\mathcal{R}^z$ by Eq. (3)
		\State Calculate adaptive task score  $\alpha^z$ by Eq. (5)
		\State Reweighting $\mathcal{R}^z$ by Eq. (6)
		\State Calculate $\mathcal{P}^q$ by Eq. (7) and Eq. (8)
		\EndFor
		\State  $L$ $\leftarrow$ $-\sum \mathcal{Y}log(\mathcal{P})$
		\State 
		mini-Batch Adam to minimize $L$, update $\theta$, $\phi$, $\varphi$ and $\omega$
		\EndWhile
	\end{algorithmic}
\end{algorithm}

	\noindent
	\textbf{Implementation of MATANet}
	
	We provide a PyTorch implementation of MATANet for few-shot learning. Our code is avaliable at \href{https://github.com/chenhaoxing/MATANet}{https://github.com/chenhaoxing/MATANet}.
	
\end{document}